\documentclass[10pt,journal,compsoc]{IEEEtran}
\usepackage{lmodern}
\usepackage[pdftex]{graphicx}
\usepackage{amsmath,amssymb} % define this before the line numbering.
\usepackage{color}
\usepackage{epsfig}
\usepackage{epstopdf}
\usepackage{amssymb}
\usepackage{subfig}
\usepackage{gensymb}
\usepackage{enumerate}
\usepackage{url}

\ifCLASSOPTIONcompsoc
  \usepackage[nocompress]{cite}
\else
  \usepackage{cite}
\fi

\ifCLASSINFOpdf
\else
\fi

\begin{document}
\title{A Simple, Fast and Highly-Accurate Algorithm to Recover 3D Shape from 2D Landmarks on a Single Image}

\author{Ruiqi Zhao, Yan Wang, Aleix M. Martinez% <-this % stops a space
\IEEEcompsocitemizethanks{\IEEEcompsocthanksitem RZ, YW and AMM are with the Department
of Electrical and Computer Engineering, The Ohio State University, Columbus,
OH, 43210.\protect\\
http://cbcsl.ece.ohio-state.edu.}% <-this % stops a space
\thanks{Manuscript received April 19, 2005; revised August 26, 2015.}}


\markboth{Journal of \LaTeX\ Class Files,~Vol.~14, No.~8, August~2015}%
{Shell \MakeLowercase{\textit{et al.}}: Bare Advanced Demo of IEEEtran.cls for IEEE Computer Society Journals}



\IEEEtitleabstractindextext{%
\begin{abstract}
Three-dimensional shape reconstruction of 2D landmark points on a single image is a hallmark of human vision, but is a task that has been proven difficult for computer vision algorithms. We define a feed-forward deep neural network algorithm that can reconstruct 3D shapes from 2D landmark points almost perfectly (i.e., with extremely small reconstruction errors), even when these 2D landmarks are from a single image. Our experimental results show an improvement of up to two-fold over state-of-the-art computer vision algorithms; 3D shape reconstruction of human faces is given at a reconstruction error $<.004$, cars at $.0022$, human bodies at $.022$, and highly-deformable flags at an error of $.0004$. Our algorithm was also a top performer at the 2016 3D Face Alignment in the Wild Challenge competition (done in conjunction with the European Conference on Computer Vision, ECCV) that required the reconstruction of 3D face shape from a single image.  The derived algorithm can be trained in a couple hours and testing runs at more than $1,000$ frames/s on an i7 desktop. We also present an innovative data augmentation approach that allows us to train the system efficiently with small number of samples. And the system is robust to noise (e.g., imprecise landmark points) and missing data (e.g., occluded or undetected landmark points).
\end{abstract}

}

\maketitle

% is not selected <OR> if conference mode is selected - because compsoc
% conference papers position the abstract like regular (non-compsoc)
\IEEEdisplaynontitleabstractindextext

% \fi
\IEEEpeerreviewmaketitle

\section{Introduction}
\label{sec:introduction}

\IEEEPARstart{H}{umans} can readily and accurately estimate the 3D shape of an object from a set of 2D landmark points on a single image. Many computer vision approaches have been developed over the years in an attempt to replicate this outstanding ability.  As an example, the approaches in \cite{Zhou_2015_CVPR,ramakrishna2012,lin2014jointly,categoryShapesKar15,HamsiciGM12} provide 3D shape estimates of a set of 2D landmark points on a single image. Other works, such as structure from motion (SfM) and shape from shading, require a sequence of images \cite{fayad2011automated,6126319,Jeni:15}, and although they have been extensively studied, the results are not as accurate as the ones reported in the present paper. 

The above mentioned algorithms learn a set of 3D shape bases, with the assumption that the deformation in a test image can be represented as a linear combination of these bases. This linear assumption limits the applicability of these approaches on highly deformable or articulated shapes, a point that is only exacerbated when the number of 2D landmark points is small (i.e., non-dense fiducials). To resolve these problems, some previous works, e.g., \cite{ramakrishna2012,lin2014jointly}, are specifically designed to recover the 3D shape of a specific object by introducing domain-specific priors into the 3D estimate. However, the addition of these priors typically limit the applicability of the resulting algorithms, e.g., to improve generic object recognition \cite{lin2014jointly,Leotta2011} or estimate 3D pose in virtual reality \cite{Lee1985,ramakrishna2012}. Another solution is to identify the set of picewise segments that belong to the same surface and then solve for each section separately \cite{Russell:11,Russell:14}. However, these algorithms require multiple images or video sequences to identify consistency and smoothness in the movement.

Estimating the 3D geometry of an object from a single view is an ill-posed problem. Nevertheless, with the available 3D ground-truth of a number of 2D sample images with corresponding 2D landmarks, we can learn the mapping from 2D to 3D landmark points, i.e., we can define a mapping function $f(.)$ that given a set of 2D landmark points $x$ outputs their corresponding 3D coordinates $y$, $y=f(x)$. There are three main challenges in this task: (1) how to define a general algorithm that aims at a wide range of rigid {\em and} non-rigid objects; (2) how to define an algorithm that yields good results whether we have a small or a large number of samples; (3) how to make this process run in real-time ($>60$ frames/s).

\begin{figure*}[htp]
\centering{
\includegraphics[width=7in]{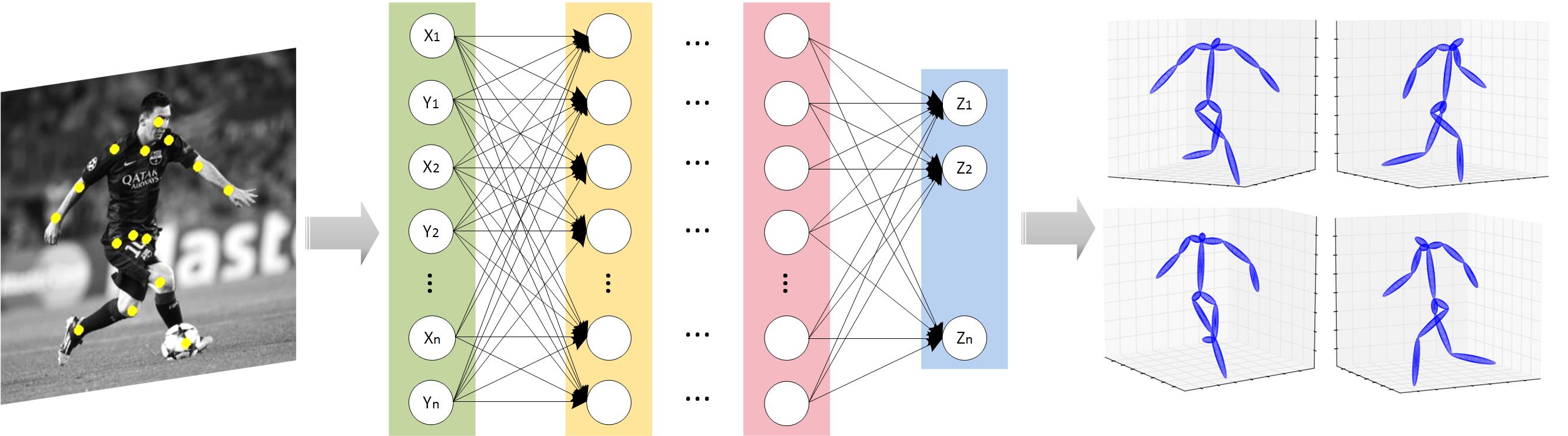}
\caption{Framework of the proposed algorithm to estimate the 3D shape of a set of 2D landmark point on a single image. The resulting 3D shape is estimated in real-time ($>1,000$ frames/s on an i7 desktop).}\label{fig:example}
}
\end{figure*}

In order to deal with the aforementioned challenges, we propose a deep-network framework to estimate the function $f(.)$. The proposed model is illustrated in Figure~\ref{fig:example}. 

Unlike most existing methods, our work derives a deep neural network that uses a set of hierarchical nonlinear transformations to recover depth information of a number of known 2D landmark points on a single image. The 3D shape is estimated by combining the input and output of the neural network and is independent of any scaling factor. The algorithm can be efficiently trained regardless of the number of samples and is robust to noise and missing data. This derived deep neural network is efficient, outperforming  previous algorithms by a significant margin. 
%............ models introduces an augmented shape-space model, \cite{ramakrishna2012} presents an activity-independent method to reconstruct human pose from 2D image landmarks, \cite{Lin2014} employs a deformable 3D model for landmark locations of car images, \cite{categoryShapesKar15} learns 3D shape models from silhouettes and model deformations by linear combination of bases. Other works such as structure from motion \cite{} are also  

The major contributions of the herein derived algorithm can be summarized as follows.
\begin{enumerate}[(1)]

\item Our algorithm is extremely general and can be applied to recover the 3D shape of basically any rigid or non-rigid object for which a small or large number of training samples (i.e., 2D and corresponding 3D landmark points) is available. As examples, we provide results on human faces, cars, human bodies and flags. 

\item  Our algorithm is not limited by the number of training samples.  When the number of training data is very large, we employ mini-batch training to parallelize gradient-based optimization and reduce computational cost. When the training set is small, we used an innovative data augmentation approach that can generate many novel samples of the 2D landmark points of a given set of 3D points using several camera models; this yields new 2D landmark points as viewed from multiple cameras, scales, points of view, etc.

\item Our proposed multilayer neural network can be trained very quickly. Additionally, in testing, our algorithm runs much faster than real-time ($> 1,000$ frames/s on an i7).
\end{enumerate}

\section{Related Work}

Estimating the 3D geometry of an object from a single view is an ill-posed problem. There has been substantial work on detecting 2D landmark points \cite{Rivera:12,DelaTorre:13,Yang:2011} from a single image, but how about 3D shape? Recently, a number of databases including accurate 3D and 2D landmark points of different objects have allowed the learning of mapping functions between 2D and 3D. Example databases are the Google 3D Warehouse \cite{WinNT}, the Carnegie Mellon Motion capture set \cite{WinNT}, the Fine-Grained 3D Car database (FG3DCar) \cite{lin2014jointly}, and the PASCAL3D database \cite{Aggapito_PAMI:16,xiang_wacv14}.

Directly related to our work are several approaches that reconstruct 3D shape from a single image \cite{Zhou_2015_CVPR,ramakrishna2012,lin2014jointly,vicente2014reconstructing,categoryShapesKar15}. Previous methods usually fit a shape-space model to recover the 3D configuration of the object from its 2D locations of landmarks in a single image \cite{ramakrishna2012,Zhu2015ICCV,Zhou_2015_CVPR,ZhouZLD15}. In \cite{ramakrishna2012}, the authors address the human pose recovery problem by presenting an activity-independent method, given 2D locations of anatomical landmarks. Unfortunately, this method is limited to the reconstruction of human bodies. Lin \emph{et al.} \cite{lin2014jointly} derived an algorithm to recover 3D models of cars. In the work of \cite{Zhu2015ICCV}, an optimization method for location analysis is proposed to map the discriminant parts of the object into a 3D geometric representation. In \cite{Zhou_2015_CVPR}, a shape-based model is designed, which represents a 3D shape as a linear combination of shape bases.  In \cite{ZhouZLD15}, the authors extend this previous work to deal with outliers in the 2D correspondences and rotation asynchronization. More broadly, but also related to our work, reconstructing 3D shape from a sequence of 2D images using structure from motion (SfM) has been extensively investigated \cite{Hartley:2003}. In particular, the work of \cite{HamsiciGM12} can recover the 3D shape of an objects from 2D landmark points on a single image using SfM-trained models.

These works assume a low-dimensional embedding of the underlying shape space, and represent each 3D shape as a linear combination of shape bases. For example, Principal Component Analysis (PCA) and other linear methods are employed to obtain the low-dimensional shape bases \cite{Gotardo_PAMI,Akhter:11}. A major limitation of linear models is that when the true (but unknown) mapping function is nonlinear, the accuracy drops significantly \cite{Gotardp_ICCV}. Therefore, these methods cannot efficiently handle highly deformable or articulated shapes. 

Additionally, several algorithms are limited to a certain type of object categories, e.g. \cite{ramakrishna2012} is designed for reconstructing 3D human pose from 2D image landmarks and \cite{lin2014jointly} focuses on 3D car modeling. These algorithms thus make prior assumptions or use special geometric properties of the object to constrain the solution space that do not typically generalize well to other object categories. 

An additional limitation of the above cited papers is their inability to learn from either large or small datasets. Some existing algorithms require very small training sets, but are unable to yield improvements when larger datasets are available. On the other hand, some algorithms require very large training sets and are incapable of yielding reasonable results when the number of samples is small.

Our theoretical and experimental results reported below demonstrate how the herein derived algorithm resolves the limitations of the methods discussed in this section.

\section{Proposed Approach}

In this section, we describe how to recover the 3D shape of an object given its 2D landmarks on a single image. 

\subsection{Preliminaries}

Let us denote the $n$ 2D landmark points on the  $i^{th}$ image 
\begin{equation}
{\bf W}_i= \left( {\begin{array}{*{20}{c}}
{\begin{array}{*{20}{c}}
{{u_{i1}}}&{{u_{i2}}}& \cdots &{{u_{in}}}
\end{array}}\\
{\begin{array}{*{20}{c}}
{{v_{i1}}}&{{v_{i2}}}& \cdots &{{v_{in}}}
\end{array}}
\end{array}} \right)\in\mathbb{R}^{2\times n},
\end{equation}
where $(u_{ij},v_{ij})^T$ are the 2D coordinates of the $j^{th}$ image landmark point. Our goal is to recover the 3D coordinates of these 2D landmark points,
\begin{equation}
{\bf S}_i=\left( {\begin{array}{*{20}{c}}
{{x_{i1}}}&{{x_{i2}}}& \cdots &{{x_{in}}}\\
{{y_{i1}}}&{{y_{i2}}}& \cdots &{{y_{in}}}\\
{{z_{i1}}}&{{z_{i2}}}& \cdots &{{z_{in}}}
\end{array}} \right)\in\mathbb{R}^{3\times n},
\end{equation}
where $(x_{ij},y_{ij},z_{ij})^T$ are the 3D coordinates of the landmarks of the $j^{th}$ object. 

Assuming a weak-perspective camera model, with calibrated camera matrix $\mathcal{M}=\begin{pmatrix}
\lambda &0  &0 \\ 
0 &\lambda  &0 
\end{pmatrix}$, the weak-perspective projection of the object 3D landmark points is given by
\begin{equation}
\label{eq:linear}
{\mathbf W}_i = \mathcal{M}{\bf S}_i.
\end{equation}
This result is of course defined up to scale, since $\boldsymbol{u}_i=\lambda \boldsymbol{x}_i$ and $\boldsymbol{v}_i=\lambda \boldsymbol{y}_i$,  where ${\mathbf x}_i^T=(x_{i1},x_{i2},...,x_{in})$, ${\mathbf y}_i^T=(y_{i1},y_{i2},...,y_{in})$, ${\mathbf z}_i^T=(z_{i1},z_{i2},...,z_{in})$, ${\mathbf u}_i^T=(u_{i1},u_{i2},...,u_{in})$ and ${\mathbf v}_i^T=(v_{i1},v_{i2},...,v_{in})$. This will require that we standardize our variables when deriving our algorithm.

\subsection{Deep 3D Shape Reconstruction from 2D Landmarks}

\subsubsection{Proposed Neural Network.}\label{Sec: Proposed NN}
Given a training set with $m$ 3D landmark points $\{{\bf S}_i\}_{i=1}^{m}$, we aim to learn the function $f: {\mathbb{R}}^{2n} \to {\mathbb{R}}^{n}$, that is,
\begin{equation}
\widehat{{\bf z}}_{i}=f({\widehat{{\bf x}}_{i}}, {\widehat{{\bf y}}_{i}}),
\end{equation}
where $\widehat{{\bf x}}_{i}$, $\widehat{{\bf y}}_{i}$ and $\widehat{{\bf z}}_{i}$ are obtained by standardizing ${\bf x}_i$, ${\bf y}_i$ and ${\bf z}_i$ as follows, 
\begin{equation}
\begin{aligned}
\widehat{x}_{ij}=\frac{x_{ij}-\overline{{\bf x}}_i}{(\sigma({\bf x}_i)+\sigma({\bf y}_i))/2},\\
\widehat{y}_{ij}=\frac{y_{ij}-\overline{{\bf y}}_i}{(\sigma({\bf x}_i)+\sigma({\bf y}_i))/2},\\
\widehat{z}_{ij}=\frac{z_{ij}-\overline{{\bf z}}_i}{(\sigma({\bf x}_i)+\sigma({\bf y}_i))/2},\\
\end{aligned}
\end{equation}
where $\overline{{\bf x}}_i$, $\overline{{\bf y}}_i$ and $\overline{{\bf z}}_i$ are mean values, and $\sigma({\bf x}_i)$, $\sigma({\bf y}_i)$ and $\sigma({\bf z}_i)$ are the standard deviation of the elements in ${{\bf x}_i}$, ${{\bf y}_i}$ and ${{\bf z}_i}$, respectively. 

We standardize ${\bf x}_i$, ${\bf y}_i$ and ${\bf z}_i$ to eliminate the effect of scaling and translation of the 3D object, as noted above. By estimating $f$ using a training set, we learn the geometric property of the given object. Herein, we model the function $f(.)$ using a multilayer neural network. 

Figure~\ref{fig:example} depicts the overall architecture of our neural network. It contains $L$ layers. The $l^{th}$ layer is defined as,
\begin{equation}
\boldsymbol{a}^{(l+1)}=\tanh\left({\bf\Omega}^{(l)}\boldsymbol{a}^{(l)}+\boldsymbol{b}^{(l)}\right),
\end{equation}
where $\boldsymbol{a}^{(l)}\in\mathbb{R}^{d}$ is an input vector, $\boldsymbol{a}^{(l+1)}\in\mathbb{R}^{r}$ is the output vector, $d$ and $r$ specify the number of input and output nodes, respectively, and ${\bf\Omega}\in\mathbb{R}^{r\times d}$ and $\boldsymbol{b}\in\mathbb{R}^{r}$ are network parameters, with the former a weighting matrix and the latter a basis vector. Our neural network uses a Hyperbolic Tangent function, $\tanh(\cdot)$ .

Our objective is to minimize the sum of Euclidean distances between the predicted depth location $\boldsymbol{a}_i^{(L)}$ and the ground-truth $\widehat{\boldsymbol{z}}_i$ of our $m$ 3D landmark points. Formally,
\begin{equation}
\displaystyle \min{\sum_{i=1}^m{\| \widehat{\boldsymbol{z}}_i-\boldsymbol{a}^{(L)}_i \|_2}},
\end{equation}
with $\| . \|_2$ the Euclidean metric. 

We utilize the RMSProp \cite{Tieleman2012} algorithm to optimize our model parameters. In a multilayer neural network, the appropriate learning rates can vary widely during learning and between different parameters. RMSProp is a technique that updates parameters of a neural network to improve learning. It can adaptively adjust the learning rate of each parameter separately to improve convergence to a solution. 

\subsubsection{Testing.}
When testing on the $t^{th}$ object, we have ${\bf u}_t$ and ${\bf v}_t$, and want to estimate ${\bf x}_t$, ${\bf y}_t$ and ${\bf z}_t$ . From Eq.~(\ref{eq:linear}) we have $\boldsymbol{u}_t=\lambda \boldsymbol{x}_t$ and $\boldsymbol{v}_t=\lambda \boldsymbol{y}_t$. Thus, we first standardize the data,
\begin{equation} \label{eq:std}
\begin{aligned}
\widehat{u}_{tj}=\frac{u_{tj}-\overline{\boldsymbol{u}}_t}{(\sigma(\boldsymbol{u}_t)+\sigma(\boldsymbol{v}_t))/2},\\
\widehat{v}_{tj}=\frac{v_{tj}-\overline{\boldsymbol{v}}_t}{(\sigma(\boldsymbol{u}_t)+\sigma(\boldsymbol{v}_t))/2}.
\end{aligned}
\end{equation}
This yields $\widehat{\boldsymbol{x}}_t=\widehat{\boldsymbol{u}}_t$ and $\widehat{\boldsymbol{y}}_t=\widehat{\boldsymbol{v}}_t$. This means we can directly feed $({\widehat{\boldsymbol{u}}_{t}}, {\widehat{\boldsymbol{v}}_{t}})$ into the trained neural network to obtain its depth $\widehat{\boldsymbol{z}}_{t}$. Then, the 3D shape of the object can be recovered as $(\widehat{\boldsymbol{u}}_{t}^T, \widehat{\boldsymbol{v}}_{t}^T, \widehat{\boldsymbol{z}}_{t}^T)^T$, a result that is defined up to scale.

\subsubsection{Missing Data.}

\begin{figure}[htp]
\centering{
\includegraphics[width=2in]{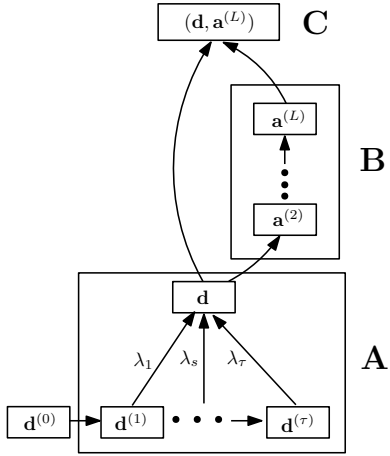}
\caption{Deep network for dealing with missing data. ${\mathbf d}^{(0)}$ is the input to the network, ``A" is the recurrent layer with $\tau$ steps for estimating missing inputs. ``B" is the multi-layer neural network defined in Section \ref{Sec: Proposed NN} and summarized in Figure~\ref{fig:example}. ``C" combines the results of ``A" and ``B" to yield the final output of the network.}\label{fig:rnn}
}
\end{figure}

To solve the problem of missing data, we  add a recurrent layer \cite{bengio1996recurrent} on top of the multi-layer neural network to jointly estimate both the 2D coordinates of missing 2D landmarks and their depth. This is illustrated in Figure~\ref{fig:rnn}. The module named ``A" corresponds to the recurrent layer that estimates the 2D entries of the missing data, while ``B" is the multi-layer neural network described before and summarized in Figure~\ref{fig:example}. The output of ``A" is thus the full set of 2D landmarks and the output of ``B" their corresponding depth values. The module ``C" merges the outputs of ``A" and ``B" to generate the final output, $\left( \widehat{\boldsymbol{u}}_{i}^T, \widehat{\boldsymbol{v}}_{i}^T, \widehat{\boldsymbol{z}}_{i}^T \right)^T$, and  ${\ell _2}$ is the loss function used in this module. 

In the recurrent network, we use the notation ${\widehat u}_{ij}^{(s)}$ and ${\widehat v}_{ij}^{(s)}$ to specify the estimated values of ${\widehat u}_{ij}$ and ${\widehat v}_{ij}$ at iteration $s$. Here $i$ represents the $i^{th}$ sample. The input to our above defined network can then be written as ${\mathbf d_i}^{(0)}=({\widehat u}_{i1}^{(0)}, {\widehat v}_{i1}^{(0)}, \cdots , {\widehat u}_{in}^{(0)}, {\widehat v}_{in}^{(0)})$, with $s=0$ specifying the initial input. If the values of $u_{ij}$ and $v_{ij}$ are missing (i.e., occluded in the image), then ${\widehat u}_{ij}^{(0)}$ and ${\widehat v}_{ij}^{(0)}$ are set to zero. Otherwise the values of $u_{ij}$ and $v_{ij}$ are standardized using Eq. \eqref{eq:std} to obtain ${\widehat u}_{ij}^{(0)}$ and ${\widehat v}_{ij}^{(0)}$.

In subsequent iterations, from $s-1$ to $s$, if the $j^{th}$ landmark is not missing, ${\widehat u}_{ij}^{(s)}={\widehat u}_{ij}^{(s-1)}$ and ${\widehat v}_{ij}^{(s)}={\widehat v}_{ij}^{(s-1)}$. If the $j^{th}$ landmark is missing, then ${\widehat u}_{ij}^{(s)}=g \left( \sum\limits_{k = 1}^{2n} {{w_{k(2j - 1)}}d_{ik}^{(s - 1)}} \right)$, ${\widehat v}_{ij}^{(s)}=g \left( \sum\limits_{k = 1}^{2n} {{w_{k(2j)}}d_{ik}^{(s - 1)}} \right)$, where $g( \cdot )$ can be the identity function or a nonlinear function (e.g. $\tanh(\cdot)$) and $w_{k(2j-1)}$, $w_{k(2j)}$, $k=1,...,2n$, $j=1,...,n$ are the parameters of the recurrent layer. In our experiments we find that $g( \cdot )$ being the identity function works best. 

The number of iterations is set to $\tau$, which yields ${\mathbf d_i} = \sum\limits_{s = 1}^\tau  {{\lambda _s}{{\mathbf d_i}^{(s)}}}$, where $0 < {\lambda _1} <  \cdots  < {\lambda _\tau }$ and $\sum\limits_{s = 1}^\tau {\lambda_s} = 1$, as the final output of the recurrent layer. The vector ${\boldsymbol \lambda}=(\lambda_1,...,\lambda_{\tau})^T$ is fixed by hand. By using the weighted sum of the output at each step rather than the output at the last step as final output of the recurrent layer, we can enforce intermediate supervision to make the recurrent layer gradually converge to the correct output. 

\subsubsection{Data augmentation approach.}
In many applications the number of training samples (i.e., 2D and corresponding 3D landmark points) is small. However, any regressor designed to learn the mapping function $f(.)$ requires of a large number of training samples with the 2D landmarks as seen from as many cameras, views (translation, rotation) and scales as possible. We resolve this with a simple, yet efficient data augmentation approach.

The key to our approach is to note that, for a given object, its 3D structure does not change. What changes are the 2D coordinates of the landmark points in the image. For example, scaling or rotating an object in 3D yields different 2D coordinates of the same object landmarks. Thus, our task is to generate as many of these possible sample views as possible. We do this with a camera model. 

A camera model allows us to predict the 2D image coordinates of 3D landmark points. Here, we use an affine camera model to generate a very large number of images of the known 3D sample objects. We do this by varying the intrinsic parameters of the camera model (e.g., focal length) as well as the extrinsic parameters (e.g., 3D translation, rotation and scale). Specifically, we use the weak-perspective camera model defined above.  

We also use this data augmentation step to model imprecisely localized 2D landmark points. All detection algorithms include a detection error (even when fiducial detections are done by humans) \cite{Ding:10}. We address this problem by modeling the detection error as Gaussian noise, with zero mean and variance $\sigma$. Specifically, we use a small variance equivalent to about 3\% of the size of the object. This means that, in addition to the 2D landmark points given by the camera models used above, we will incorporate 2D landmark points that have been altered by adding this random Gaussian noise. This allows our neural network to learn to accurately recover the 3D shape of an object from imprecisely localized 2D landmark points.

It is important to note that, when the original training set is small, we can still train an efficiently-performing neural network using this trick. In fact, we have found experimentally that we do not need a large number of training samples to obtain extremely low reconstruction errors using our derived approach and this data augmentation trick. When the number of samples is large, this approach can help reduce the 3D reconstruction error by incorporating intrinsic or extrinsic camera parameters and detection errors not well represented in the samples.

\subsubsection{Implementation Details.}
Our feed-forward neural network contains six layers. The number of nodes in each layer is $[2n,2n,2n,2n,2n,n]$.
We divide our training data into a training and a validation set. In each of these two sets, we perform data augmentation. 

We use Keras library \cite{chollet2015} on top of Theano \cite{bastien2012theano} to implement our proposed multilayer neural network. Early stopping is enabled to prevent overfitting. We stop the training process if the validation error does not decrease after $10$ iterations. We set the initial learning rate at $.01$. 

\section{Experimental Results}

We conduct experiments on a variety of databases to test the effectiveness of our algorithm. We used the following datasets: the CMU Motion Capture database \cite{CMUmotion}, the fine-grained 3D car (FG3DCar) database \cite{lin2014jointly}, the Binghamton-Pittsburgh 4D Spontaneous Expression (BU$-$3DFE) database \cite{zhang2014bp4d,zhang2013high} and the flag flapping in the wind database \cite{white2007capturing}. We also report our results on the sequestered dataset of the 3D Face Alignment in the Wild Challenge (3DFAW), done in conjunction with the 2016 European Conference on Computer Vision (ECCV), where the herein derived algorithm was a top performer.

Comparisons with the state-of-the-art demonstrates that our algorithm is significantly more accurate and efficient in recovering 3D shape from a single 2D image than previously defined methods. The 3D reconstruction error is evaluated using the Procrustes distance between the reconstructed shape and the ground-truth. Specifically, the ground-truth is given in millimeters (mm) which is normalized to a standard scale (given by the mean of all 3D landmark points) to make the error measure invariant to scale. 

Also, to demonstrate the robustness of our method, we perform sensitivity analysis on these databases. Results show that our method is tolerant to moderate Gaussian noise. We also show results of how our method handles missing data.

Our derived neural network runs much faster than real-time, $>1000$ frames/s, on a 3.40 GHz Intel Core i7 desktop computer.

\subsection{CMU Motion Capture Database}
\setlength{\tabcolsep}{4pt}
\begin{table}
\begin{center}
\caption{Comparisons of the reconstruction error on CMU Motion Capture Database. The proposed results yields a highly-significant reduction on the reconstruction error compared with previously defined algorithms.}
\label{Table:headings}
\begin{tabular}{lccc}
\hline\noalign{\smallskip}
Methods & Subject 13 & Subject 14 & Subject 15\\
\noalign{\smallskip}
\hline
\noalign{\smallskip}
This paper  & $\boldsymbol{.0229}$ & $\boldsymbol{.0201}$ & $\boldsymbol{.0099}$\\
Zhou \emph{et al.} \cite{Zhou_2015_CVPR} & $.0653$ & $.0643$ & $.0405$\\
Ramakrishna \emph{et al.} \cite{ramakrishna2012} & $.0983$ & $.0979$ & $ .0675$\\
\hline
\end{tabular}
\end{center}
\end{table}
\setlength{\tabcolsep}{1.4pt}

\begin{figure*}[htp]
\centering{
\includegraphics[width=7in]{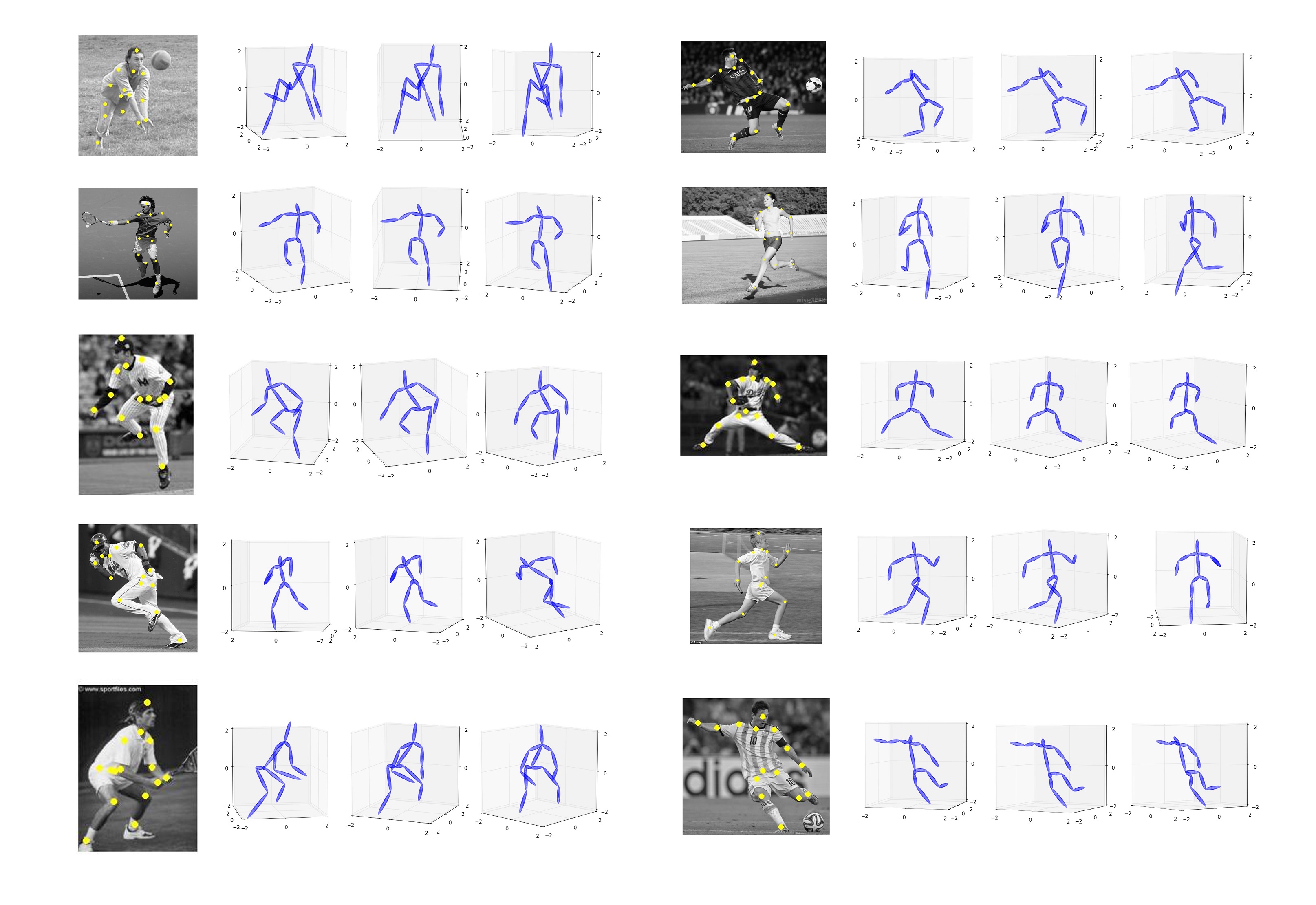}
\caption{Result of our algorithm on images of the humans in-the-wild dataset \cite{NIPS2006_2976}. The first and fifth columns are the single 2D images and 2D landmarks (in yellow) used by our algorithm. The reconstructed 3D shapes are shown on the second, third, fourth, sixth, seventh and eight columns; these show the 3D reconstruction from multiple views to more clearly demonstrate the quality of the recovered 3D shape.}
\label{Figure:example}
}
\end{figure*}

The CMU motion capture database contains 3D human body joints locations of subjects performing various physical activities. Each body shape is defined by 15 3D landmark points. To provide a fair comparison, we follow the experimental setting of \cite{Zhou_2015_CVPR}, where sequences of subject 86 are used for training and sequences of subjects 13, 14 and 15 are used for testing. During the training of the neural network, we split data of subject 86 into five folds and use four folds for training and the other fold for validation. We train the neural network in 10,000 epochs. We train one batch for at most 300 iterations within one epoch. The batch is either the original training data (the first epoch) or a random rotation of the original training data in 3D space. To represent real human shape in a 2D image, we  set the rotation angle to be uniformly distributed within the range of [-20\degree, 20\degree] about the $x$ axis, [-20\degree, 20\degree] about the $y$ axis and [-180\degree, 180\degree] about the $z$ axis. 

Comparative results are in Table~\ref{Table:headings}. As shown in this table, our results are significantly more accurate than those given by previously defined algorithms. To further demosntrate the effectiveness and generality of our method, we randomly selected several 2D images from the human-in-the-wild dataset \cite{NIPS2006_2976} and used the herein derived algorithm to recover the 3D shape of the human bodies in these images. The results are in Figure~\ref{Figure:example}. 

\subsection{3D Face Reconstruction}
\setlength{\tabcolsep}{4pt}
\begin{table}
\begin{center}
\caption{Comparisons results of our method and Zhou \emph{et al.} \cite{Zhou_2015_CVPR} on BU$-$3DFE, 3DFAW and FG3DCar.}
\label{Table:face}
\begin{tabular}{lccc}
\hline\noalign{\smallskip}
Database & This paper & Zhou \emph{et al.} \cite{Zhou_2015_CVPR}\\
\noalign{\smallskip}
\hline
\noalign{\smallskip}
BU$-$3DFE Face & $\boldsymbol{0.0040}$ & 0.0053\\
3DFAW & $\boldsymbol{0.0042}$ & 0.0055\\
FG3DCar & $\boldsymbol{0.0022}$ & 0.0042\\
\hline
\end{tabular}
\end{center}
\end{table}
\setlength{\tabcolsep}{1.4pt}

\begin{figure*}[htp]
\centering{
\includegraphics[width=7in]{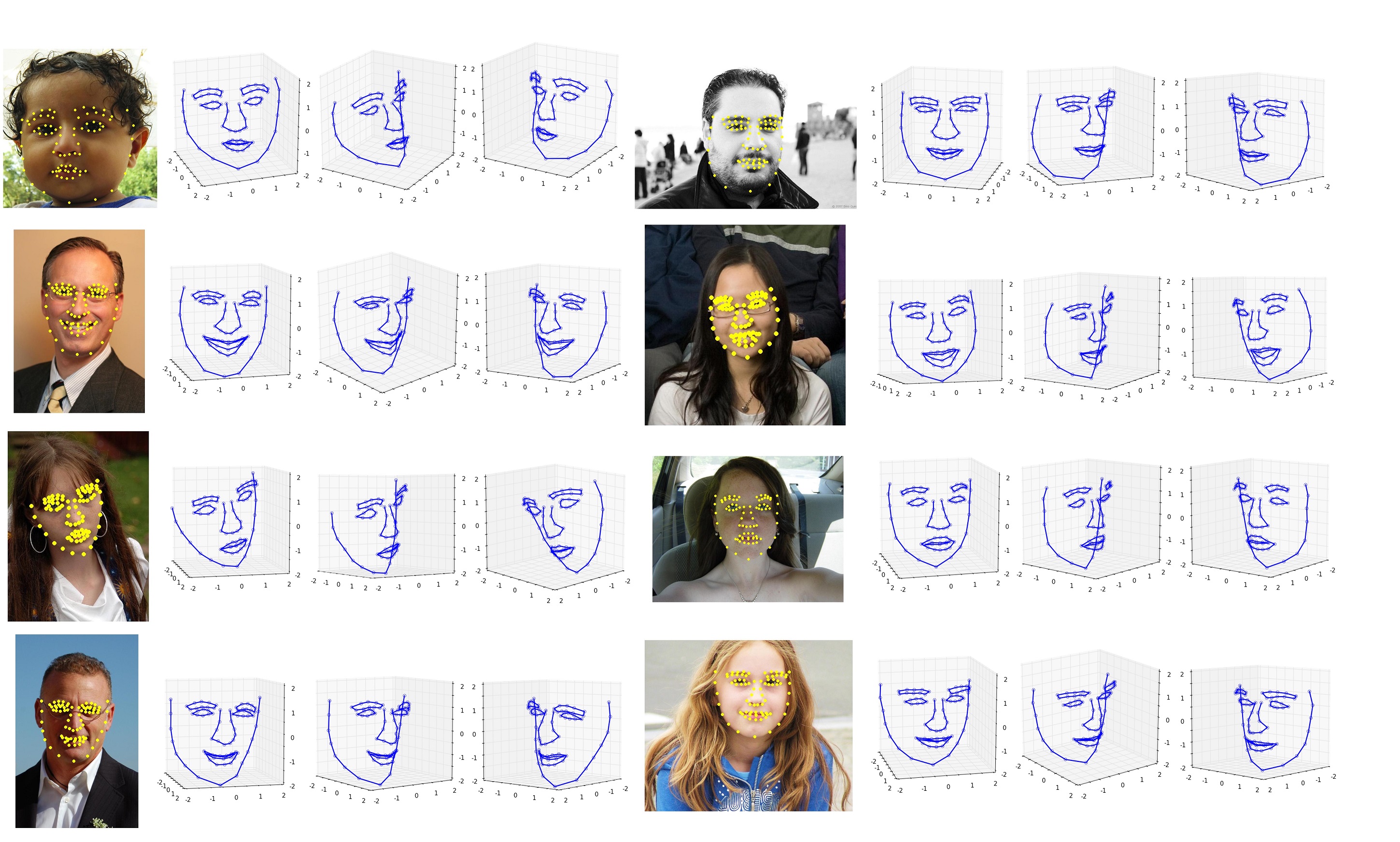}
\caption{Illustration of our 3D estimation results on the Helen database \cite{Le2012}. The model is trained on the BU$-$3DFE face database and tested on Helen. The first and fifth columns show the input 2D images with landmarks (in yellow); the other columns show the estimated 3D shapes.}
\label{Figure:face}
}
\end{figure*}

%\subsection{3D Face Alignment in the Wild Challenge Database}

The BU$-$3DFE database contains images of $100$ subjects performing six facial expressions in front of a 3D scanner. Each subject performed each expression 25 times, for a total of $2,500$ 3D sample images. Every sample has 83 annotated 3D landmarks. We randomly select 60 subjects for training, 10 for validation and the remaining 30 for testing. 

We trained our neural network in $50,000$ epochs and used the same training strategy as described in the preceding section, with the difference that we restricted rotations about the $z$ axis in the range of [-60\degree, 60\degree] to more accurately reflect real head movement. 

The mean testing error of our algorithm is ${\bf 0.004}$. To provide comparative results against the state-of-the-art, we train and test the algorithm of \cite{Zhou_2015_CVPR} and use the same experimental setting described above. Comparative results are in Table~\ref{Table:face}. These results show that the proposed algorithm outperform the state-of-the-art method of \cite{Zhou_2015_CVPR}. 

We also validate the cross database generality of our trained model on the Helen database of \cite{Le2012}. Here, we randomly select some face images and manually label the face landmarks. The reconstructed 3D shapes are shown in Figure~\ref{Figure:face}. 

To provide an additional unbiased result, we participated in the 3DFAW competition\footnote{https://competitions.codalab.org/competitions/10261}, which was conducted in conjunction with ECCV. Three of the four datasets in the challenge are subsets of MultiPIE \cite{Gross2010}, BU-4DFE \cite{Yin2008} and BP4D-Spontaneous \cite{Zhang2014} databases respectively. Another dataset TimeSlice3D contains annotated 2D images that are extracted from online videos. In total, there are $18,694$ images. Each image has 66 labeled 3D fiducial points and a face bounding box centered around the mean 2D projection of the landmarks. The 2D to 3D correspondence assumes a weak-perspective projection. The depth values have been normalized to have zero mean. Because this competition required that we also estimate the 2D landmark points in the image, we incorporated an additional deep network to our architecture to the model summarized in Figure. \ref{fig:example} and trained it to detect 2D face landmarks \cite{Zhao:16}. 

Reconstruction error was reported by the organizers of the competition on a sequestered dataset to which we did not have prior access. Our algorithm was a top performers, with a significant margin over other methods. Herein, we also provide comparative results against the algorithm of \cite{Zhou_2015_CVPR} in Table~\ref{Table:face}.

\subsection{FG3DCar Database}

\begin{figure*}[htp]
\centering{
\includegraphics[width=7in]{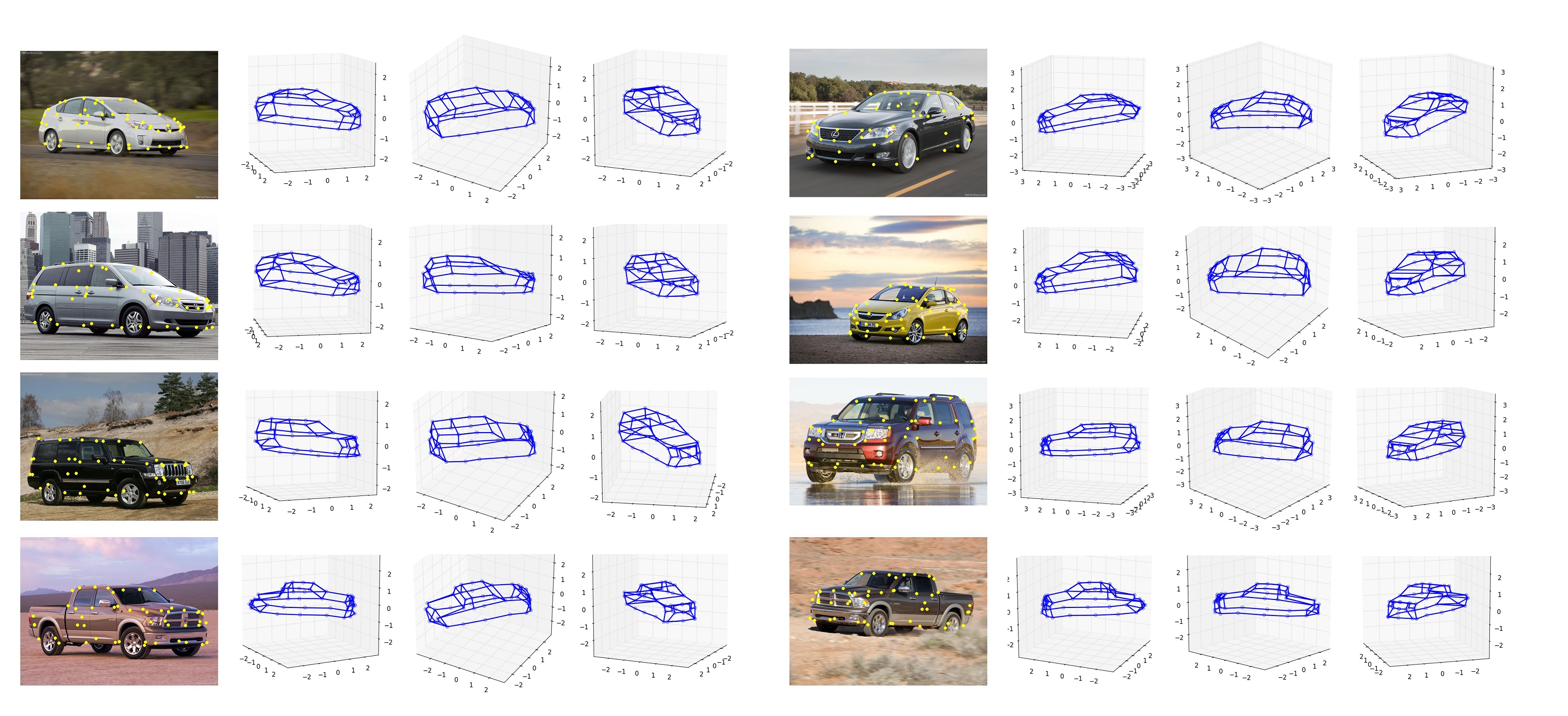}
\caption{Illustration of our 3D estimation results on the FG3DCar database. The first and fifth columns corresponds to the input 2D images and landmark points (in yellow), with the other columns showing the recovered 3D shapes.}
\label{Figure:car}
}
\end{figure*}

This is a fine-grained 3D car database. It consists of 300 images of 30 different car models as seen from multiple views. Each car model has 10 images. Each image has 64 annotated 2D landmarks and a 3D shape reconstructed CAD (Computer-Aided Design) model. We adopt the default setting for training and testing, i.e., half of the 3D shapes of each car model is used for training, the other half is used for testing. In order to train our neural network,  we further split the $150$ 3D training sample shapes into training (120 shapes) and validation (30 shapes) sets. Testing is conducted on the remaining $150$ images. 

The neural network is trained for 100,000 epochs. We follow the same procedure used in the preceding two sections and train our neural network on one batch for at most 300 iterations in each epoch. The batch is either composed of the 120 training samples (the first epoch) or a random 3D rotation of the 120 training samples. We augment the validation set $2,000$ times using the data augmentation approach described above, resulting in a total of $60,000$ validation sample images. 

The mean 3D reconstruction error on the testing set is $\boldsymbol{0.0022}$. Comparative results with the method of \cite{Zhou_2015_CVPR} is in Table~\ref{Table:face}.  Qualitative examples of our results are shown in Figure~\ref{Figure:car}. 

\subsection{Flag Flapping in the Wind Database}

\begin{figure*}[!htb]
\begin{center}
%\begin{picture}(10,20)(100,-110) %180 is height
\includegraphics[trim={7cm 4cm 5cm 4cm},clip, scale=0.5]{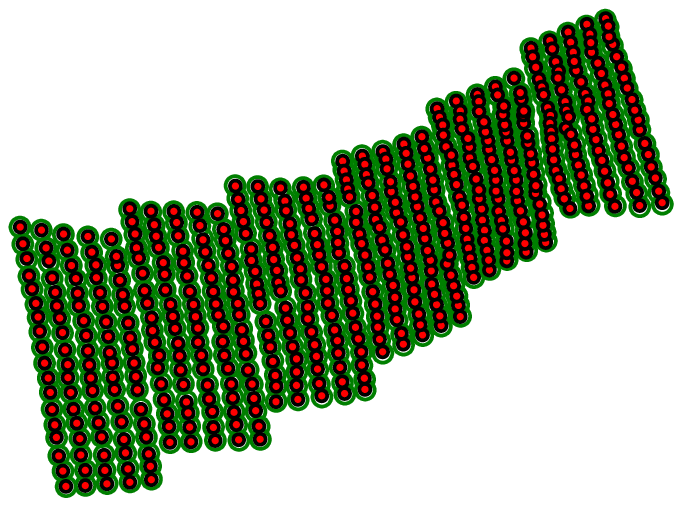}
\includegraphics[trim={7cm 4cm 5cm 4cm},clip, scale=0.5]{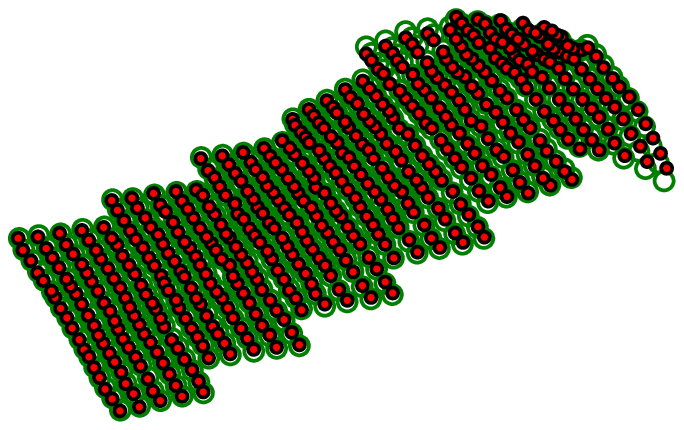}
\includegraphics[trim={7cm 4cm 5cm 4cm},clip, scale=0.5]{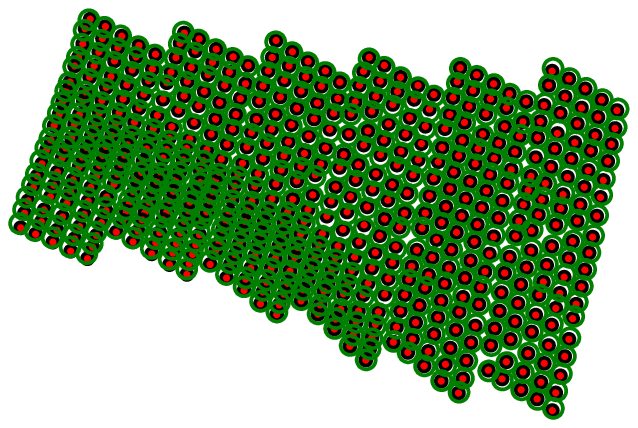}
\includegraphics[trim={7cm 4cm 5cm 4cm},clip, scale=0.5]{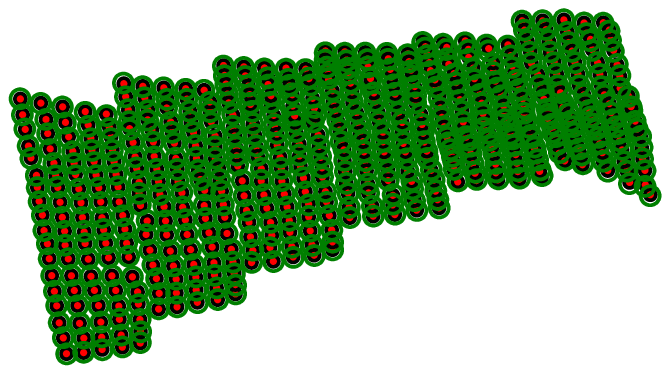} \\
\includegraphics[trim={7cm 4cm 5cm 4cm},clip, scale=0.5]{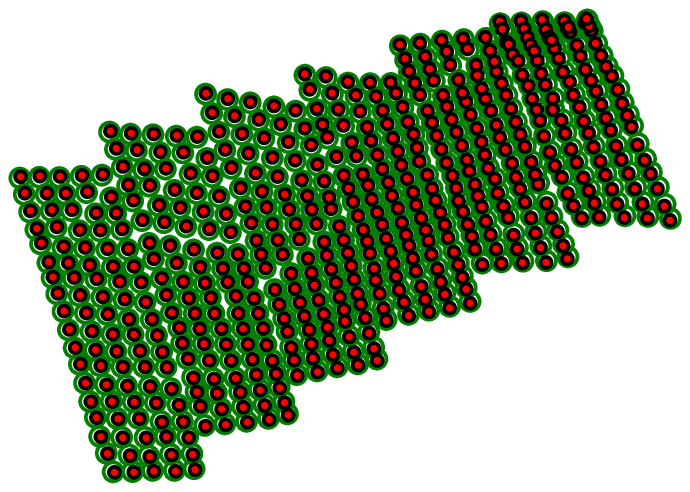}
\includegraphics[trim={7cm 4cm 5cm 4cm},clip, scale=0.5]{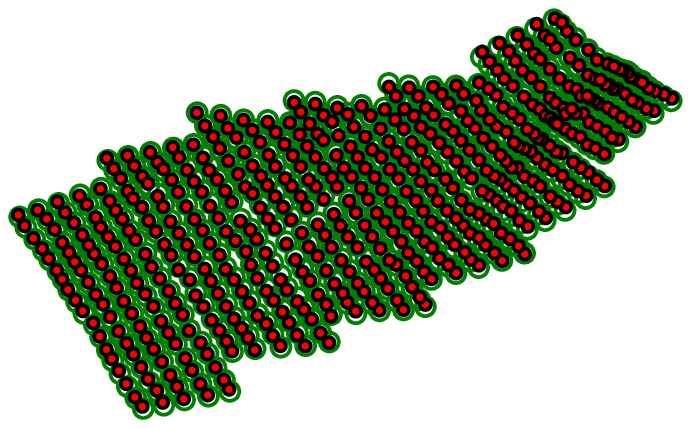}
\includegraphics[trim={7cm 4cm 5cm 4cm},clip, scale=0.5]{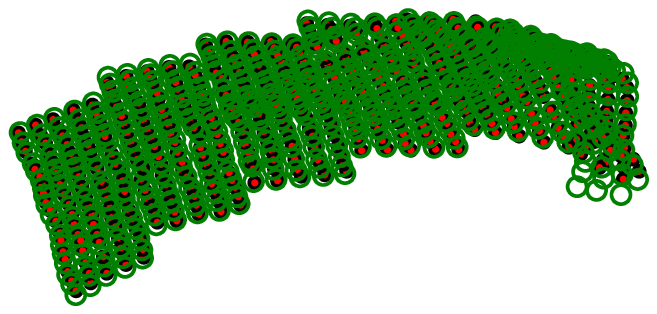}
\includegraphics[trim={7cm 4cm 5cm 4cm},clip, scale=0.5]{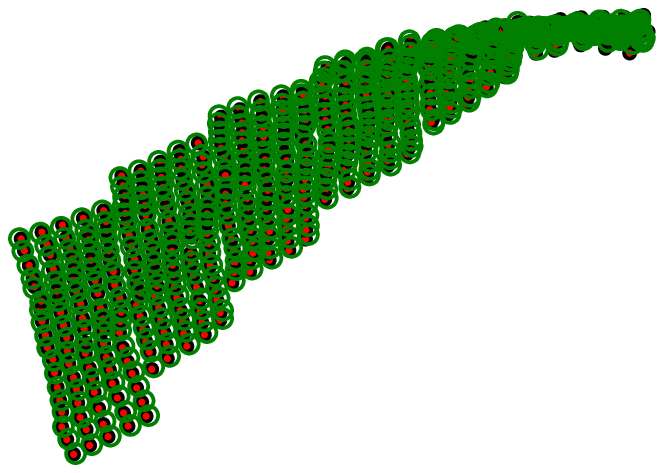}
%\end{picture}
\end{center}
%\put(-425,48){\hbox{before}}
\caption{3D shape reconstruction results on the flag flapping in the wind sequence. Comparisons of the reconstructed 3D shape (in red) with its 3D ground-truth (in green).}
\label{Figure:flag}
\end{figure*}

Flag flapping in the wind database \cite{white2007capturing} is a motion capture sequence of a flag waving in the wind. There are 450 frames, each of which has 540 vertices. We use the first 300 frames for training and the rest for testing. The network is trained in 30,000 epochs and we use the same procedure described in the preceding sections.

The mean testing error is $\boldsymbol{0.0004}$. Figure \ref{Figure:flag} shows some of our reconstructed results compared with the ground-truth. The reconstructed 3D shape is shown using filled red circles and the 3D ground-truth with open green circles. As we can see in these results, the reconstructed and true shape are almost identical.

\subsection{Noise and missing data}

\begin{figure*}[htp]
\centering{
\subfloat[\label{subfig-1:dummy}]{%
\includegraphics[trim={0cm 0cm 1cm 2.5cm},width=2.3in]{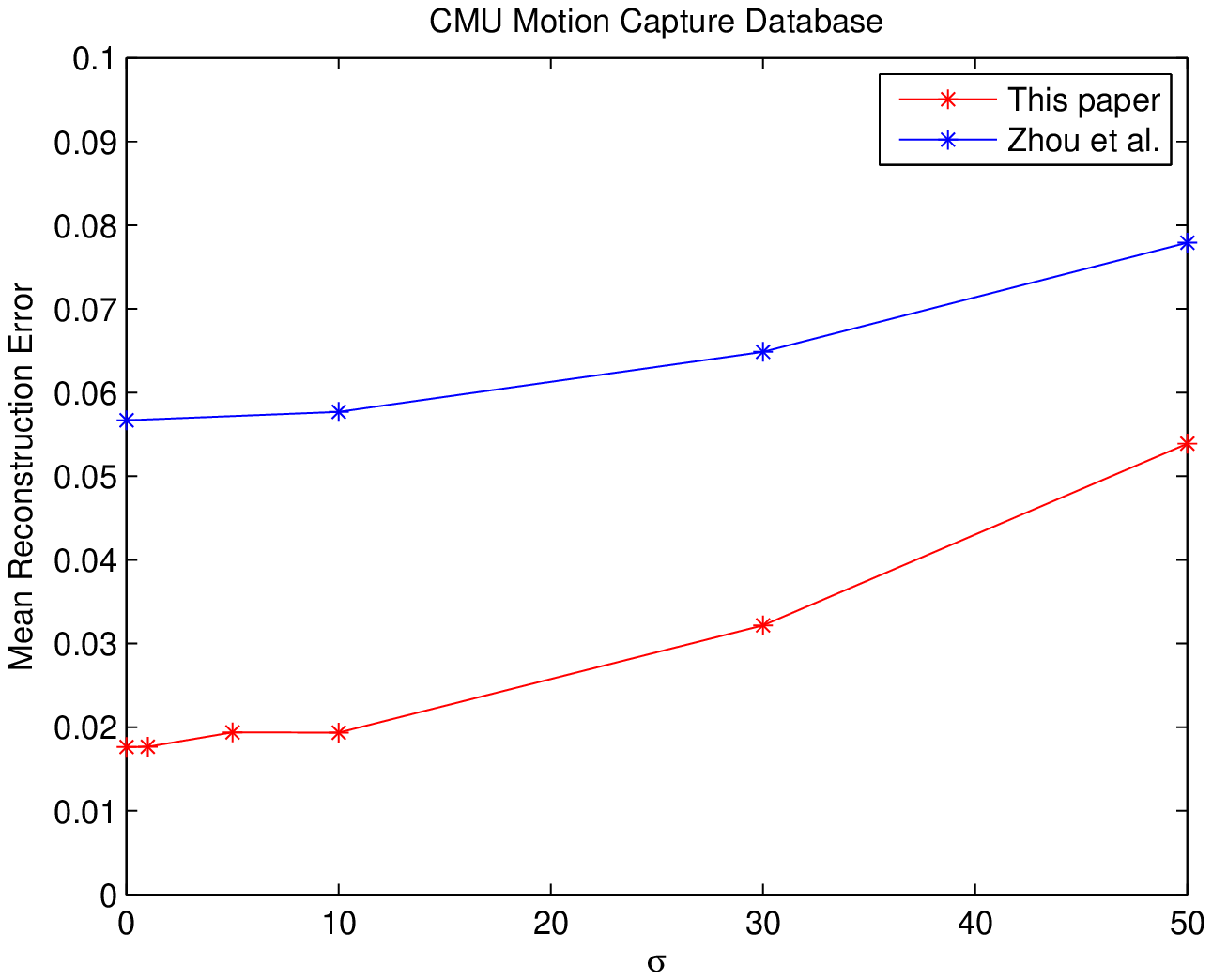}
}
\hfill
\subfloat[\label{subfig-2:dummy}]{%
\includegraphics[trim={0cm 2cm -1cm 2.5cm},clip, width=4.5in]{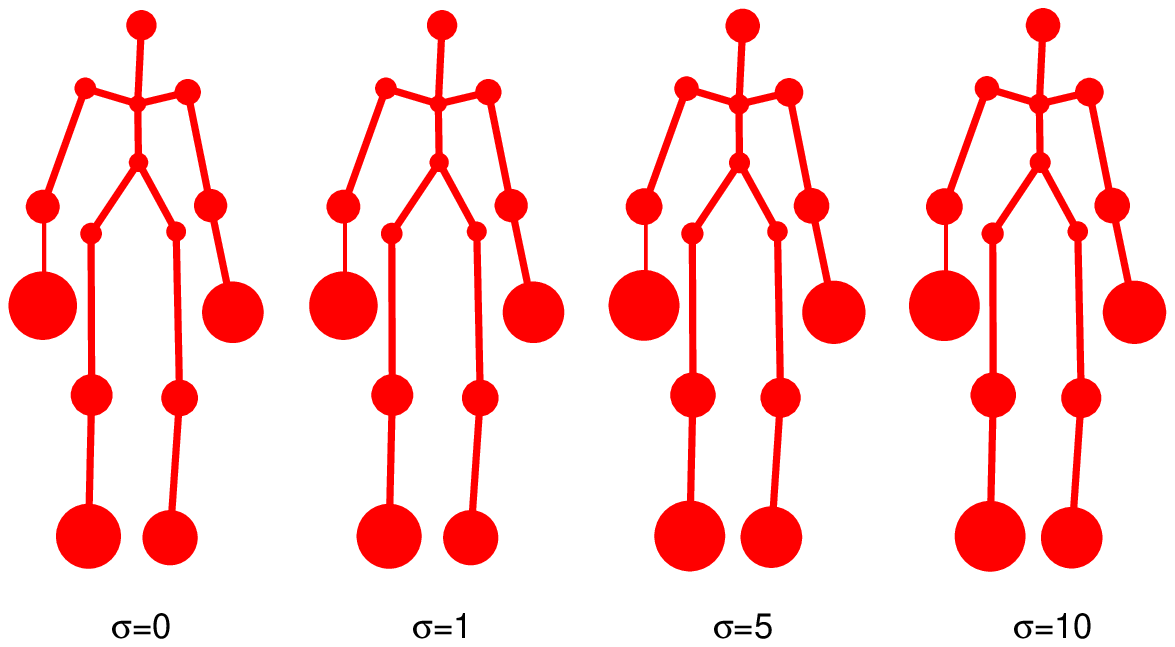}
} \\

\subfloat[\label{subfig-21:dummy}]{%
\includegraphics[trim={0cm 0cm 1cm 2cm},width=2.3in]{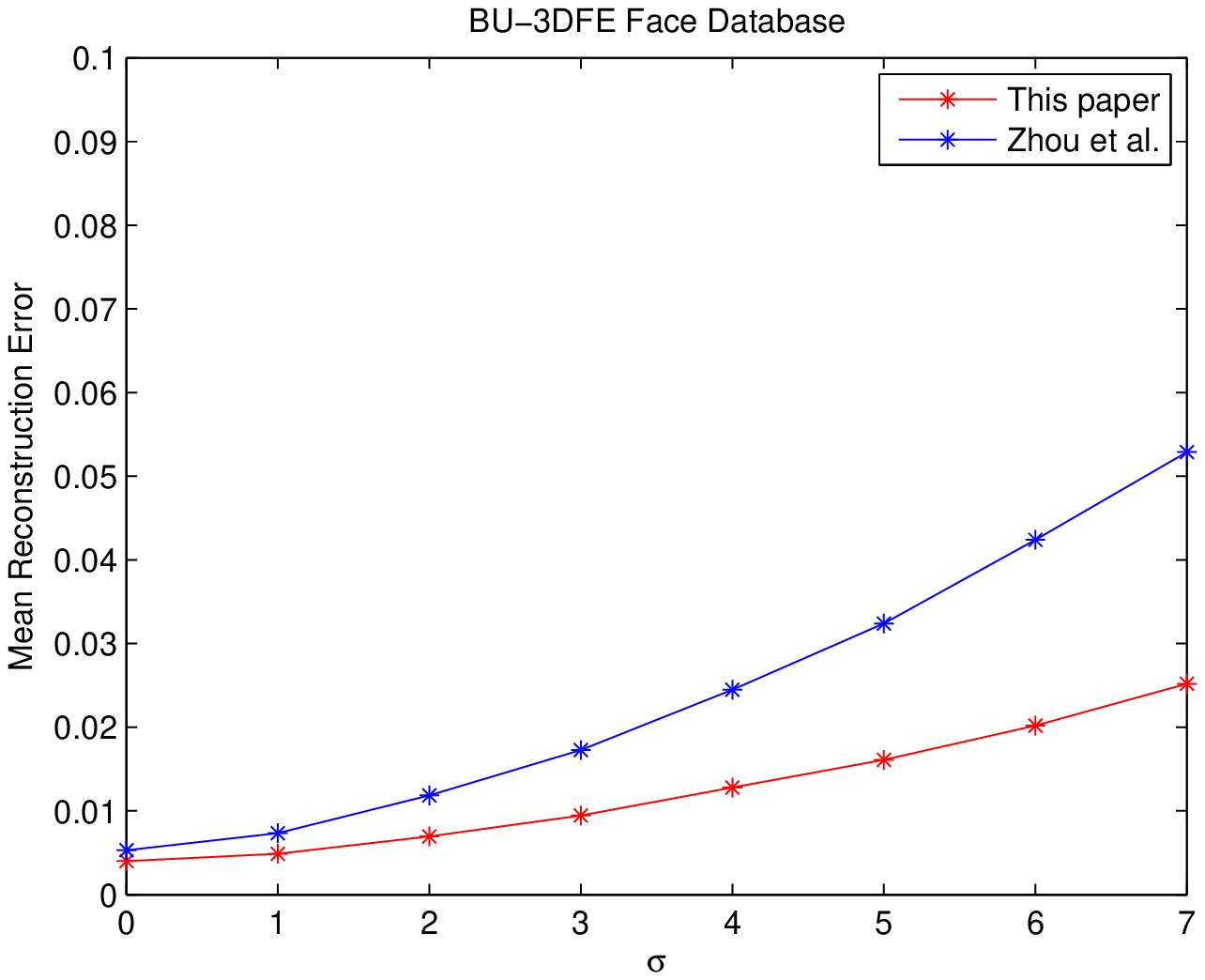}
}
\hfill
\subfloat[\label{subfig-22:dummy}]{%
\includegraphics[trim={6cm 4cm 2cm 3.5cm},clip, width=4.5in]{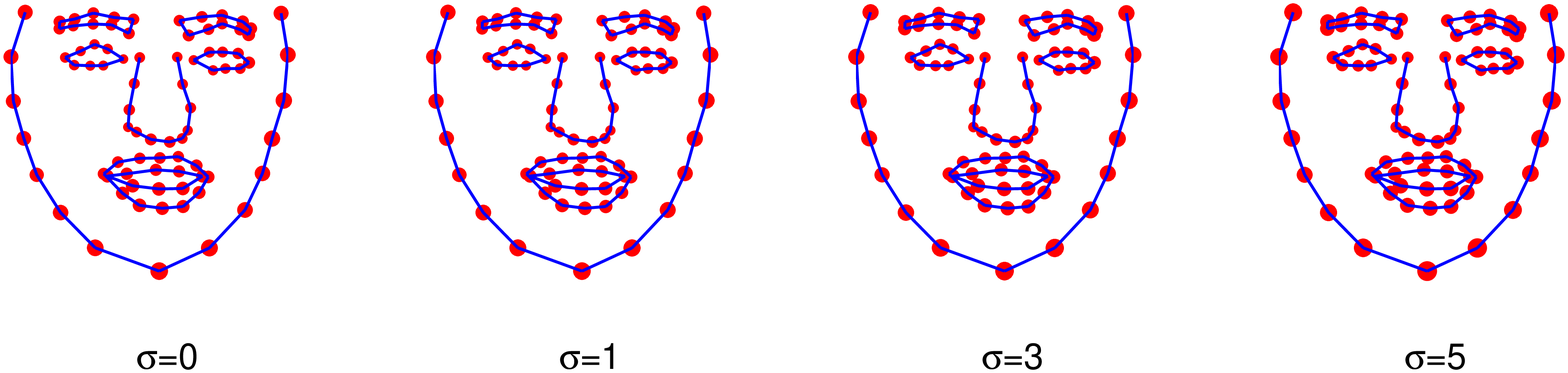}
} \\

\subfloat[\label{subfig-31:dummy}]{%
\includegraphics[trim={0cm 0cm 1cm 2cm},width=2.3in]{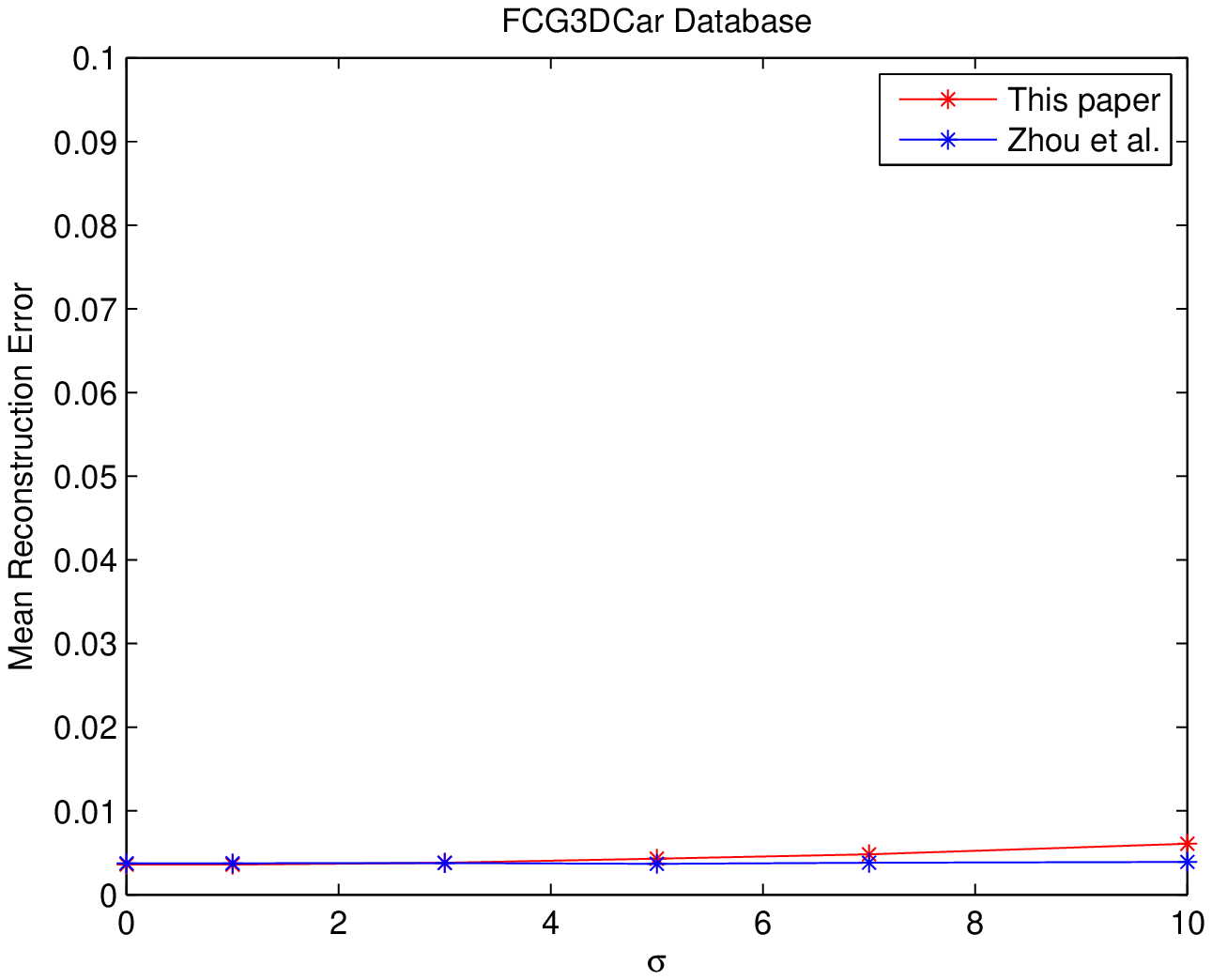}
}
\hfill
\subfloat[\label{subfig-32:dummy}]{%
\includegraphics[trim={1.6in 0in 1.5in 0in},clip, width=1.05in]{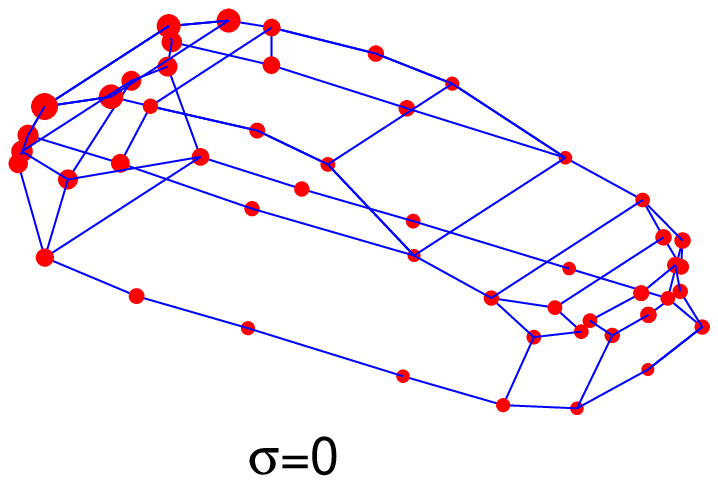}
\includegraphics[trim={1.6in -0.5in 1.45in 0in},clip, width=1.05in]{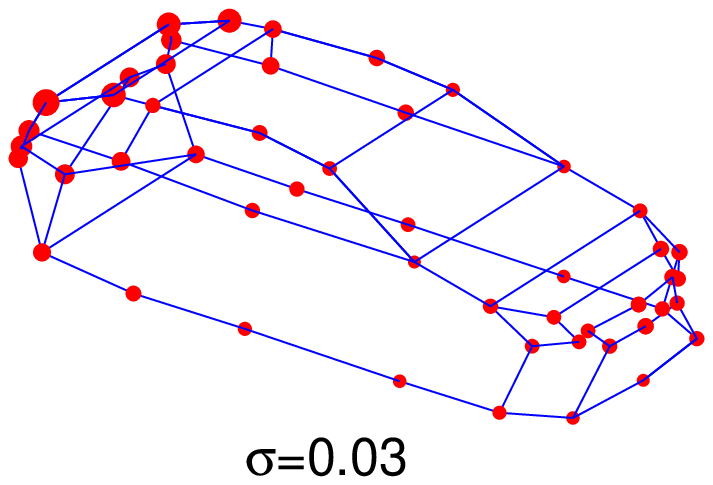}
\includegraphics[trim={2in -0.1in 0.95in 0in},clip, width=1.1in]{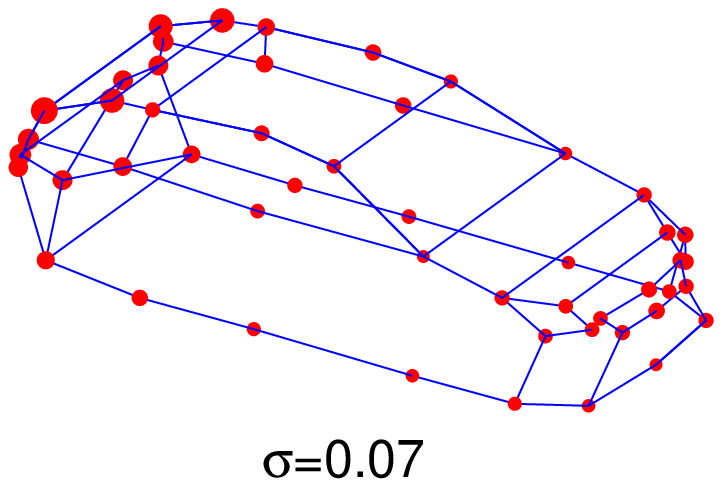}
\includegraphics[trim={1.8in -0.7in 0.8in 1.5in},clip, width=1.2in]{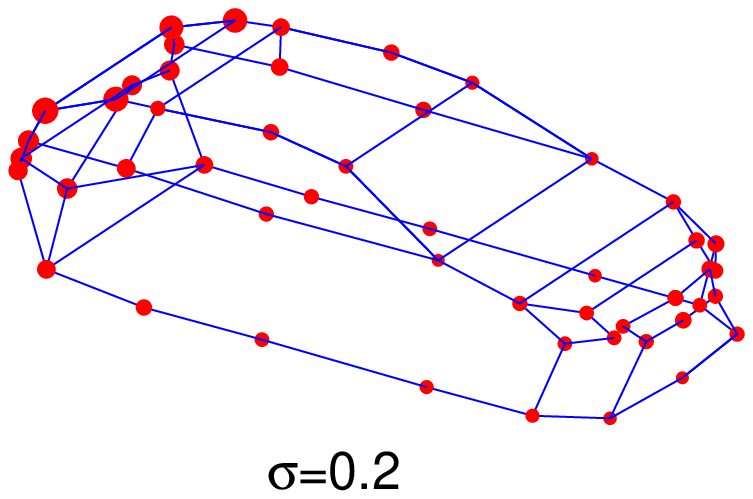}
}\\

\subfloat[\label{subfig-41:dummy}]{%
\includegraphics[trim={0cm 0cm 1cm 1cm},width=2.3in]{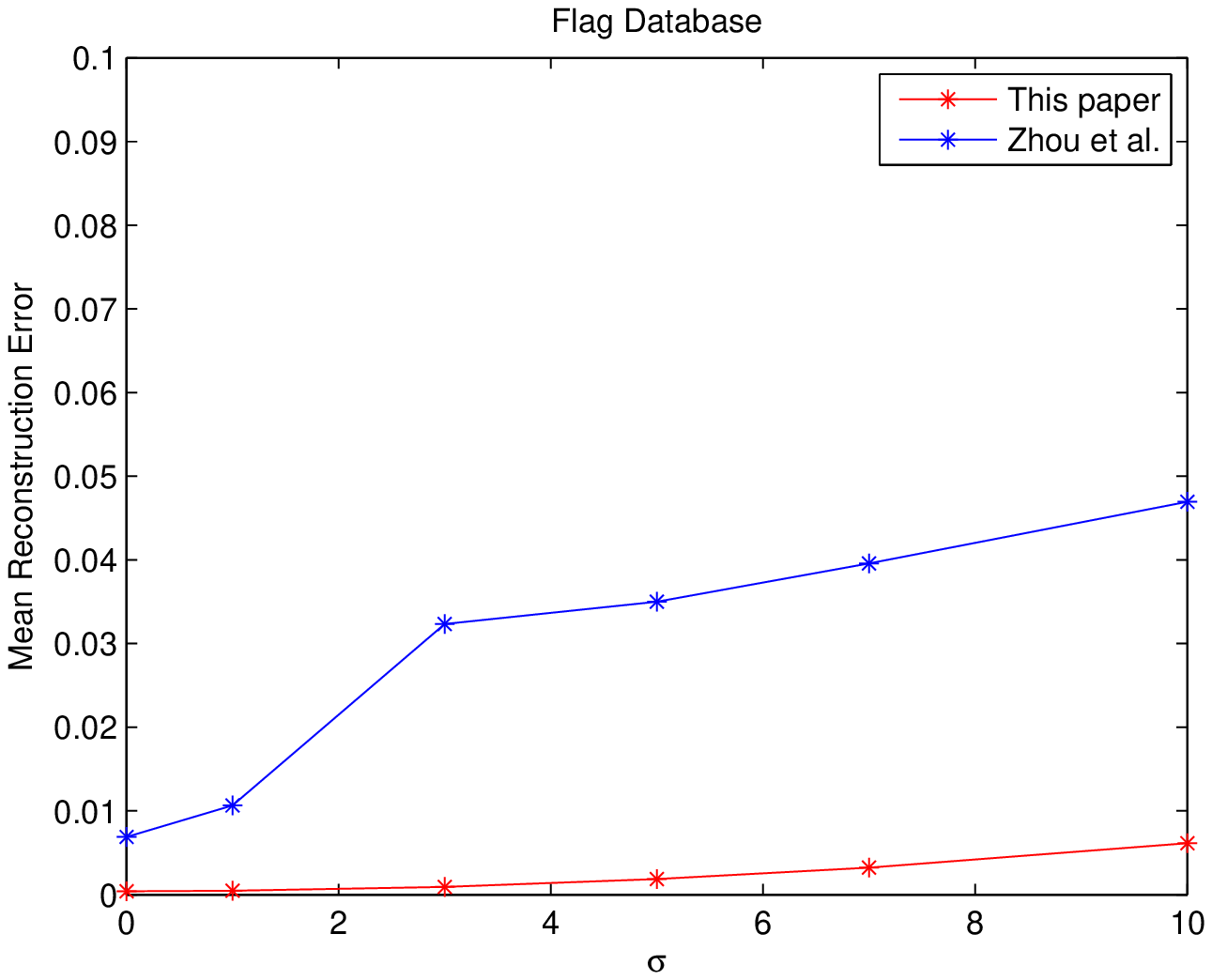}
}
\hfill
\subfloat[\label{subfig-42:dummy}]{%
\includegraphics[trim={8cm 5cm 3cm 4cm},clip, width=4.5in]{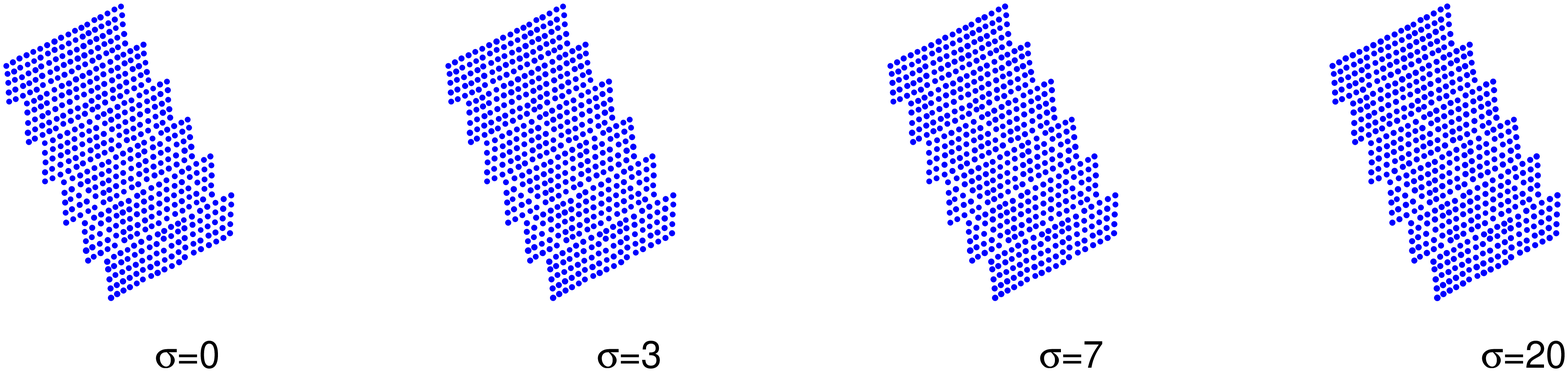}
}
\caption{Additive random Gaussian noise is used to test the robustness of the proposed algorithm to 2D inaccurate detections on CMU Motion Capture Database (a), BU$-$3DFE Database (c), FG3DCar Database (e) and Flag Flapping in the Wind Database (g). Reconstruction error is shown on the $y$ axis and $\sigma$ on the $x$ axis. Note the tinny recosntruction errors even for large values of $\sigma$. Sensitivity of reconstruction to each landmark when $\sigma$ increases on CMU Motion Capture Database (b), BU$-$3DFE Database (d), FG3DCar Database (f) and Flag Flapping in the Wind Database (h). The radius of the circle indicates the relative reconstruction error for each landmarks.}
\label{Figure:analysis_mocap}
}
\end{figure*}

To determine how sensitive the proposed neural network is to inaccurate 2D landmark detections, we add independent random Gaussian noise with variance $\sigma$ to the databases used in the preceding sections. Figure~\ref{subfig-1:dummy} shows how little the performance degrades as  $\sigma$ increases when noise is added to the CMU Motion Capture database. The average height of subjects in this dataset is $1,500$ mm, meaning the variance of the noise $\sigma$ is about $3\%$. In Figure~\ref{subfig-1:dummy}, we can see the robustness of the proposed algorithm to inaccurate 2D landmark positions. Figure~\ref{subfig-2:dummy} shows the relative reconstruction error averaged across the testing subjects for each landmark with and without noise.  

The results on the BU$-$3DFE Face Database, FG3DCar Database and Flag Flapping in the Wind sequence when the data is distorted with Gaussian noise are shown in Figures~\ref{subfig-21:dummy}-\ref{subfig-42:dummy}. The average width of the faces in BU$-$3DFE is 140 mm, hence, the variance is $5\%$.  The mean width of the car models in FG3DCar is $569$ pixel, hence, the variance is $2\%$. The mean width of the flags is $386$ mm, hence, the variance is $3\%$.

\begin{figure*}[htp]
\centering{
\subfloat[\label{subfig-human:dummy}]{
\includegraphics[width=7.2in]{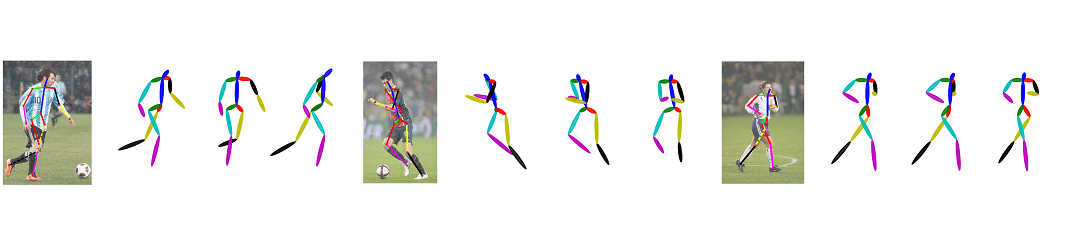}}\\
\subfloat[\label{subfig-face:dummy}]{
\includegraphics[width=7.2in]{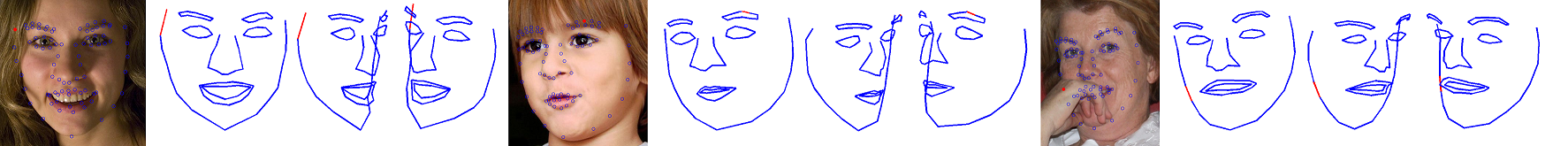}}\\
\caption{Qualitative illustration of our algorithm applied to images of human bodies and faces with missing landmark points. (a) Missing landmarks are identified with a dotted line in the images. (b) Missing landmarks are marked in red in the images.}
\label{Figure:reconstruction_missing}
}
\end{figure*}
%\subsection{Missing Data}

Finally, we tested the ability of the trained system to deal with missing data. Here, each training and validation sample had one randomly selected landmark point missing; during training and testing. For CMU Motion Capture database, the average 3D reconstruction errors for subjects 13, 14 and 15 are 0.0413, 0.0396 and 0.0307, respectively. Figure~\ref{subfig-human:dummy} shows qualitative results on three randomly selected images of humans in the wild. For BU$-$3DFE Face Database, the mean reconstruction error is 0.006. Figure~\ref{subfig-face:dummy} shows qualitative results on three randomly selected images of humans in the wild. 

\section{Conclusions}

We have presented a very simple algorithm for the reconstruction of 3D shapes from 2D landmark points that yield extremely low reconstruction errors. Specifically, we proposed to use a feed-forward neural network to learn the mapping function between a set of 2D landmark points and an object's 3D shape. The exact same neural network is used to learn the mappings of rigid (e.g., cars), articulated (e.g., human bodies), non-rigid (e.g., faces), and highly-deformable objects (e.g., flags). The system performs extremely well in all cases and yields results as much as two-fold better than previous state-of-the-art algorithms. This neural network runs much faster than real-time, $>1000$ frames/s, and can be trained with small sample sets. 

\section*{Acknowledgment}

This research was supported in part by the National Institutes of Health, grants R01-EY-020834 and R01-DC-014498, and by a Google Faculty Research Award to AMM.

\ifCLASSOPTIONcaptionsoff
  \newpage
\fi



\bibliographystyle{IEEEtran}
\bibliography{egbib_1}



%\vfill


\end{document}